\documentclass{article}
\usepackage{arxiv}

\usepackage{geometry}
\usepackage{graphicx}
\usepackage[hidelinks]{hyperref}

\usepackage{amssymb}
\usepackage{bm}
\usepackage{lineno}
\usepackage{amsmath}

\usepackage{multirow}
\usepackage{float}
\usepackage{tabularx}
\usepackage{subcaption}
\usepackage{booktabs}
\usepackage{array,xcolor}
\usepackage{arydshln}

\usepackage{algorithm, algpseudocode}

\usepackage{cite}
\title{Hybrid variable spiking graph neural networks for energy-efficient scientific machine learning}
\author{
  Isha Jain \\
  Department of Applied Mechanics\\
  Department of Mechanical Engineering\\
  Indian Institute of Technology Delhi\\
  Hauz Khas - 110 016, India.\\
  \texttt{isha.infinity162@gmail.com}
  \And
  Shailesh Garg  \\
  Department of Applied Mechanics\\
  Indian Institute of Technology Delhi\\
  Hauz Khas, New Delhi 110016, India. \\
  \texttt{shailesh.garg@am.iitd.ac.in} \\
  \And
  Shaurya Shriyam  \\
  Department of Mechanical Engineering\\ 
  Yardi School of Artificial Intelligence (YScAI)\\
  Indian Institute of Technology Delhi\\
  Hauz Khas, New Delhi 110016, India. \\
  \texttt{shriyam@mech.iitd.ac.in}\\
  \And
  Souvik Chakraborty  \\
  Department of Applied Mechanics\\
  Yardi School of Artificial Intelligence (YScAI)\\
  Indian Institute of Technology Delhi\\
  Hauz Khas, New Delhi 110016, India. \\
  \texttt{souvik@am.iitd.ac.in}}
\begin{document}
\maketitle
\begin{abstract}
Graph-based representations for samples of computational mechanics-related datasets can prove instrumental when dealing with problems like irregular domains or molecular structures of materials, etc. To effectively analyze and process such datasets, deep learning offers Graph Neural Networks (GNNs) that utilize techniques like message-passing within their architecture. The issue, however, is that as the individual graph scales and/ or GNN architecture becomes increasingly complex, the increased energy budget of the overall deep learning model makes it unsustainable and restricts its applications in applications like edge computing. To overcome this, we propose in this paper Hybrid Variable Spiking Graph Neural Networks (HVS-GNNs) that utilize Variable Spiking Neurons (VSNs) within their architecture to promote sparse communication and hence reduce the overall energy budget.
VSNs, while promoting sparse event-driven computations, also perform well for regression tasks, which are often encountered in computational mechanics applications and are the main target of this paper. Three examples dealing with prediction of mechanical properties of material based on 
microscale/ mesoscale structures are shown to test the performance of the proposed HVS-GNNs in regression tasks.
We have also compared the performance of HVS-GNN architectures with the performance of vanilla GNNs and GNNs utilizing leaky integrate and fire neurons.
The results produced show that HVS-GNNs perform well for regression tasks, all while promoting sparse communication and, hence, energy efficiency.
\end{abstract}
\section{Introduction}\label{section: introduction}
Graph Neural Networks \cite{scarselli2008graph,GNN_review,GNN} (GNNs) have gained tremendous popularity in recent years owing to their performance while dealing with datasets represented using graphs. Their applications can be found in fields related to communication networks \cite{suarez2022graph,shen2022graph,jiang2022graph}, social media \cite{GNN_socialmedia,fan2020graph,han2020graph}, medicine \cite{GNN_medicine,sun2020disease,gao2023medical}, etc. In mechanics as well, we come across problems that can benefit from GNN algorithms, for example, when dealing with irregular grids/ domains or individual particles, as in the case of Lagrangian representations, or when dealing with material properties at the microscale/ mesoscale level. Graphs provide an alternative to the matrix-based representation of datasets, and they are especially useful for cases where dependencies between different parts of a dataset's sample are complex and cannot be simplified in a grid-like manner. 

A graph consists of feature-rich nodes, and the connections between these nodes are represented using edges that may or may not have features of their own.
The key concept that enables learning in GNNs is message-passing, where the node features of the graph being learned are updated iteratively by aggregating the information received from neighboring nodes. GNNs can be trained to make predictions at the node, edge, or graph levels. Within their architecture, during message-passing and in between successive layers, the GNNs use vanilla artificial neurons \cite{ANN,jain1996artificial} to process information. Such neurons
utilize continuous activation functions that fire indiscriminately for each input, hence leading to increased computations and, thus, increased energy budget. This is unfavorable from the point of view of a sustainable future. This is further aggravated when the scale at which graph applications are leveraged calls for complex network architectures. This massive energy consumption can also hinder the applicability of GNNs in edge computing \cite{cao2020overview}.

To tackle the issue of massive energy consumption of artificial neurons, researchers in recent decades have explored the use of spiking neurons \cite{spk_neurons,ponulak2011introduction,nunes2022spiking} within deep learning architectures to conserve energy. Their motivation is drawn from biological neurons in the brain as they perform tremendous tasks with a meager energy budget, owing to the sparsity in their communication. They work on the principle of all-or-none, i.e., information is forwarded only when a certain threshold is crossed, till which point it is accumulated within the neuron. While in the literature, several spiking neuron models are available \cite{yamazaki2022spiking}, their applicability in the deep learning models is contingent on their biological plausibility and their computational complexity. For example, the Hodgkin-Huxley Neuron (HHN) model \cite{hodgkin1952quantitative,yamazaki2022spiking} is among the most biologically plausible neuron models. However, its complexity prevents its widespread use for spiking neural networks.
On the other hand, the integrate-and-fire neuron model is among the least biologically plausible neuron models as it simply integrates and then propagates the information as and when a threshold is crossed.

In this regard, the Leaky Integrate and Fire \cite{rast2010leaky,lif_neurons,yamazaki2022spiking} (LIF) neuron model has emerged as one of the more popular spiking neuron models as it is relatively easy to implement while maintaining biological plausibility. Following this, in the literature \cite{chen2024signn,yin2024continuous,li2023scaling,yamazaki2022spiking}, LIF neurons have also been used within the GNN architectures. Having said this, there exist other modifications of LIF model as well, for example Izhikevich Model \cite{izhikevich2003simple,izhikevich2007geometry,yamazaki2022spiking} approximates HHN model and provides computational efficiency similar to LIF neurons, however it requires more parameters to be tuned for effective results. Another variation on LIF is an adaptive exponential integrate-and-fire neuron model \cite{brette2005adaptive,yamazaki2022spiking} that features an adaptive threshold and exponential spike mechanism, but its biological realism increases its complexity and hence makes it more suitable for applications where biological realism is the priority. 
Although spiking neurons producing binary spikes like LIF neurons have shown great performance in classification tasks, 
extending it for regression tasks is a relatively less-explored topic  \cite{henkes2024spiking,gehrig2020event,kahana2022spiking,zhang2023artificial}. 

In this paper, our overarching goal is to develop a spiking variant of GNNs for computational mechanics that has a low associated energy budget. For any artificial neuron-based deep learning model, this can be achieved in two ways. The first is to convert a trained version of the Artificial Neural Network (ANN) to its spiking variant, and the second is to train the spiking variant of the ANN natively. The converted ANNs are known to be inefficient, and converting all types of continuous activations to their spiking variant is non-trivial \cite{eshraghian2023training,wu2019direct,zhou2024direct}. This limits the customization in the original ANN as well. Natively trained SNNs, on the other hand, are more efficient and customizable. Now, in response to the low accuracy of vanilla spiking neurons in regression tasks, Variable Spiking Neurons \cite{garg2023neuroscienceinspiredscientificmachine1} (VSNs) were proposed for data-driven and energy-efficient solutions in computational mechanics. However, the performance of a deep learning model using solely VSNs is not on par with that of a similar model using artificial neurons. Hence, in this paper, we propose natively trained Hybrid Variable Spiking Graph Neural Networks (HVS-GNNs) that utilize the recently proposed VSNs in conjunction with the vanilla artificial neurons within their architecture.

The choice of the VSNs is governed by the empirical evidence available in \cite{garg2023neuroscienceinspiredscientificmachine1,garg2024neuroscience}, which shows that their performance is better in regression tasks when compared to LIF neurons. VSN model combines the dynamics of artificial neurons and the LIF spiking neurons. This gives them the advantage of both continuous activations and sparse communication. Because of the involvement of continuous activations, the VSNs transfer information using non-binary or graded spikes. This behavior makes VSNs more expensive than vanilla spiking neurons like LIF neurons in terms of energy consumption; however, because of sparse communication, the same are more energy efficient than the current state-of-the-art artificial neurons. The evidence to support the use of graded spikes is observed in the literature, as biological neurons have been shown to utilize graded spikes \cite{GOLDING19981189,JOHNSON2023105914} for communication. Also, the neuromorphic hardware \cite{neuromorphic_hardware,young2019review,balaji2019mapping,farsa2019low} chips, which are designed to support spiking neural networks and benefit from their sparse nature, are now being developed to support graded spikes \cite{ivanov2022neuromorphic,eshraghian2023training}.

To test the performance of the proposed HVS-GNNs, we use three examples from computational mechanics covering datasets at different scales and having graph inputs and that require prediction either at a graph level or at a node level. The performance of HVS-GNNs in out-of-distribution predictions has also been reported in one of the examples. Apart from this, comparisons have been drawn with graph architectures utilizing LIF neurons and vanilla artificial neurons within their architecture. The results produced support the prospective use of HVS-GNNs for data-driven computational mechanics. The spiking activity of neurons has been reported as an indicative measure of energy consumption in various examples.

The rest of the paper is arranged as follows, section \ref{section: background} introduces the HVS-GNNs. Section \ref{section: ni} discusses various examples on computational mechanics at different scales, and section \ref{section: conclusion} concludes the paper's findings.

\section{Proposed approach}\label{section: background}
In this section, we introduce Hybrid Variable Spiking Graph Neural Networks (HVS-GNNs) and show how VSNs are assimilated within graph neural network architectures in order to benefit from their energy-efficient tendencies.
The proposed HVS-GNN is highly flexible and any existing GNN can be converted into the hybrid variable spiking variant with minimal effort. Accordingly, in this section, we first review some popular GNNs and then discuss how an existing GNN can be converted into the hybrid variable spiking variant.

\subsection{Graph neural network}
GNNs \cite{scarselli2008graph}, cater to a specific group of datasets whose samples are represented using graphs. In its simplest form, a graph $\mathcal G(\bm N, \bm E)$ is a mathematical structure consisting of a set of nodes, $\bm N \in \mathbb R^n$, that are connected via a set of edges $\bm E \in \mathbb R^e$, depicting relationships between various nodes. The edges connecting any two nodes can be directed or undirected depending on the dataset under consideration. A node $n_i\in \bm N$ of a graph $\mathcal G$ consists of a set of features $\bm F_i\in\mathbb R^{f_1}$ that define the characteristics of that node. An edge connecting $i$\textsuperscript{th} node with the $j$\textsuperscript{th} node can also have a set of features $\bm{\overline F}_{ij}\in\mathbb R^{f_2}$ associated with it, similar to node features. Labels can also be assigned to the nodes and edges of a graph.
When creating a graph dataset, the nodes and edges can be assigned certain meanings that are relevant to the problem at hand, and their features can be assigned so that they provide an adequate representation of the raw data. For example,
in a communications graph, the nodes can represent various Internet-of-Things devices, and the edges can represent physical connections between any two devices. Specifically, in computational mechanics problems, if we are representing a finite element sample as a graph, the nodes can represent the locations of grid points, and edges can represent connections between close grid points, which are highly co-related. In a molecular analysis study, the nodes can be used to represent individual atoms/ molecules/ holes, and edges can represent connections between any two such entities. There are no strict restrictions placed on the definition of nodes and edges, and they can be used to represent any quantity as per the requirement of the dataset. In the examples discussed later in the paper, we define crystal grains or atoms as nodes of the graph and connections between any two such elements are represented using edges.

Since in a graph dataset, across two different graph samples, there is no restriction placed on the number of nodes and the number of edges, representing such datasets using fixed dimension matrices is non-trivial and is impossible in most cases. Hence, in graph neural networks, message-passing is used to learn the node features. In message-passing, the feature information from neighboring nodes of the node under consideration is collected and combined with the feature information of the current node to learn the complex relationships in the graph. This operation is conducted in $L$ layers, and for an intermittent layer $l$, the operation can be defined as follows,
\begin{equation}
    \begin{gathered}
    \bm{m}_i^{(l)}=\mathcal A_{\forall j\in\mathcal N_i}\left(\mathcal U_1\left(\bm{h}_i^{(l-1)}, \bm{h}_j^{(l-1)},\bm{\overline F}_{ij}\right)\right),\,l = 1,\dots,L,\,\bm{h}_i^0 = \bm F_i,\\
     \bm{h}_i^{(l)}=\mathcal U_2\left(\bm{h}_i^{(l-1)}, \bm{m}_i^{(l)}\right),
    \end{gathered}
\end{equation}
where $\mathcal A$ is an permutation invariant aggregator operator, $\bm h_i^{(l)}$ is the feature vector for $i$\textsuperscript{th} node in layer $l$, and $\mathcal N_i$ is the set of nodes, neighboring to the $i$\textsuperscript{th} node. $\mathcal U_1(\cdot)$ and $\mathcal U_2(\cdot)$ are the message and update functions, respectively, and these can be simple linear neural network layers wrapped in activation functions. The message and update functions selected must be differentiable. The aggregator operator $\mathcal A$ should be selected such that it can take any number of input values and is differentiable so as to facilitate backpropagation. Some examples of common aggregator operators used in GNNs include mean operator, sum operator, min operator, max operator, etc.

Apart from message-passing layers, GNNs can also employ linear layers wrapped in activation functions to increase or decrease the dimensionality of the feature space. Similarly, recurrent layers can also be used, which work in the feature space. Pooling layers can be used for cases where the GNNs are expected to work at a graph level and produce outputs that have fixed dimensions, such as scalar outputs. These layers collapse the graph along the feature space. An example of such a layer is the global mean pooling layer that generates output as follows,
\begin{equation}
    \bm{\tilde o} = \dfrac{1}{n} \sum_{i=1}^n \bm h_i^{(L)}.
\end{equation}
These layers allow the user to deploy linear layers wrapped in activation functions as successive layers and generate a final output of the desired dimension. In the first example discussed in the following section, we use the global mean pool layer to produce scalar outputs given graph inputs.
\subsubsection{Graph convolution network}
An important variant of GNN, often used in literature, is the Graph Convolution Network \cite{kipf2016semi,zhang2019graph} (GCN).
GCNs utilize graph convolution layers that generalize the convolution operation for the graph datasets. Before discussing message-passing in graph convolution layers, we need to introduce some helper matrices constructed using the graph's nodes and edges information. 
The first matrix is an adjacency matrix $\mathbf A\in \mathbb R^{n\times n}$, which has $n$ rows and $n$ columns corresponding to $n$ nodes of the graph.  The entry in the $i$\textsuperscript{th} row and $j$\textsuperscript{th} column of the adjacency matrix, $A_{ij}=1$, if $i$\textsuperscript{th} node is connected to the $j$\textsuperscript{th} node, else its value is set to zero. The adjacency matrix can be a weighted matrix as well, where the non-zero values can be used to denote the relative strength of the connection between any two nodes. Another useful matrix is a degree matrix $\mathbf D \in \mathbb R^{n\times n}$. It is a diagonal matrix where the entries in the diagonal show the number of edges that are connected to any particular node. The number corresponding to the $i$\textsuperscript{th} node gives the degree of that node. Finally, a node feature matrix $\mathbf N_f \in \mathbb R^{n\times f}$ is defined, whose $i$\textsuperscript{th} row contains the $f$ features corresponding to $i$\textsuperscript{th} node. Now, among $L$ message-passing layers, the operation in layer $l$ of GCN can be defined as,
\begin{equation}
\mathbf H^{(l)} = \sigma(\mathbf D^{-1/2}\tilde{\mathbf A}\mathbf D^{-1/2}\mathbf H^{(l-1)}\mathbf W^{(l)}),\,l = 1,\dots,L,\,\mathbf H^{(0)} = \mathbf N_f,
\end{equation}
where $\sigma(\cdot)$ is an activation function, $\mathbf H^{(l)}$ is the feature matrix after $l$ message-passing layers and $\mathbf W^{(l)}\in\mathbb R^{f\times f}$ is a weight matrix with learnable parameters for layer $l$. $\tilde{\mathbf A} = \mathbf A+\mathbb I$, where $\mathbb I\in \mathbb R^{n\times n}$ is an identity matrix.
In the literature, there exist other variants of GCN layers, each with unique benefits. Here, we discuss Principal Neighborhood Aggregation Convolution \cite{corso2020principal} (PNAConv) and Sample and Aggregator Convolution (SAGEConv) \cite{hamilton2017inductive} layers, which are also used in the following numerical illustrations. The PNAConv layer uses multiple aggregate functions and degree scalers to combine the incoming information from the neighboring nodes. The operation of PNAConv in $l$\textsuperscript{th} layer, among $L$ message-passing layers is defined as,
\begin{equation}
\begin{gathered}
    \bm h_i^{(l)} = \mathcal V_1 \left( \bm h_i^{(l-1)}, \underset{j \in \mathcal N_i}{\bigoplus} \mathcal V_2 \left( \bm h_i^{(l-1)}, \bm h_j^{(l-1)}, \bm{\overline F}_{ij} \right) \right)\\[0.5em]
    \bigoplus = \begin{bmatrix}
    \mathbf I \\
    S(\mathbf{D}, \alpha=1) \\
    S(\mathbf{D}, \alpha=-1)\end{bmatrix}\otimes \begin{bmatrix}
    \mathcal A_1 \\
    \mathcal A_2 \\
    \vdots \\
    \mathcal A_n
    \end{bmatrix},
\end{gathered}    
\end{equation}
where $\mathcal V_1$ and $\mathcal V_2$ are neural networks, $\mathcal A_i$ are different aggregator functions, $S(\cdot)$ is a scaler and $\alpha$ is its variable parameter. An example of a logarithmic scaler function $S(d,\alpha),\,d\in\mathbf D,\,d>0$ is defined as,
\begin{equation}
    S(d,\alpha) = \left(\dfrac{\log (d+1)}{\delta}\right)^\alpha,\, -1<\alpha<1,
\end{equation}
where $\delta$ is a normalization parameter and is taken equal to average degree of all nodes in the training set. Similarly, the operations of SAGEConv layer is defined as,
\begin{equation}
    \bm h_i^{(l)} = \mathbf{W}_1^{(l)} \bm h_i^{(l-1)} + \mathbf{W}_2^{(l)} \mathcal A_{\forall j \in \mathcal N_i} (\bm h_j^{(l-1)}),
\end{equation}
where $\mathbf W_1^{(l)}\in \mathbb R^{f\times f}$ and $\mathbf W_2^{(l)}\in \mathbb R^{f\times f}$ are the weight matrices to be learned. The dimension of the feature space in GNNs can change after each message-passing layer, and the same is be governed by the choice of update functions, message functions, and the dimensions of the weight matrices. 
\subsection{Hybrid variable spiking graph neural network}
GNN architectures utilize continuous activations at various locations. They may be present within the message-passing layers, or they may be used between two successive layers.
Linear/ nonlinear continuous activation functions fire continuously irrespective of the input, increasing their neuron activity and consequently their energy expenditure.
%
The key idea in HVS-GNN is to  strategically place variable spiking neuron \cite{garg2023neuroscienceinspiredscientificmachine1} (VSN) at selected location so as to minimize the energy consumption. 
VSNs produce graded spikes that allow for sparse event-driven processing while retaining information-rich communication. This sparsity in communication and, consequently, computations leads to a decreased energy budget. The VSN combines the dynamics of LIF neurons (refer \ref{appendix:LIF}) and artificial neurons (refer \ref{appendix:LIF}), and the output for spiking neuron at $\bar t$ Spike Time Step\footnote{Spike Time Step (STS): Information in spiking neurons is passed in the form of spike trains and the STS refers to a particular time step of input spike train to any spiking neuron.} (STS) can be calculated as follows,
\begin{equation}
    \begin{aligned} & M^{(\bar{t})}=\beta M^{(\bar{t}-1)}+z^{(\bar{t})}, \\ & \tilde{y}^{(\bar{t})}=\left\{\begin{array}{ll}1 ; & M^{(\bar{t})} \geq T_h \\ 0 ; & M^{(\bar{t})}<T_h\end{array} \quad \text {, if } \tilde{y}^{(\bar{t})}=1, M^{(\bar{t})} \leftarrow 0\right. \\ & y^{(\bar{t})}=\sigma\left(z^{(\bar{t})} \tilde{y}^{(\bar{t})}\right), \quad \text { given, } \sigma(0)=0 \text {, }  & \end{aligned}
\end{equation}
where $M^{(\bar{t})}$ and $z^{(\bar{t})}$ represent the memory and input to VSN neuron, respectively, at $\bar{t}$ STS. The memory of a VSN neuron is multiplied by a leakage parameter $\beta$ after each STS. $y^{(\bar{t})}$ represents the output at $\bar{t}$ STS and is non-zero only if $M^{(\bar{t})}$ is greater then the threshold $T_h$ of neuron. Otherwise, no output is forwarded from the neuron. The continuous activation $\sigma(\cdot)$ in VSN is selected such that $\sigma(0) = 0$. This is done to ensure that no information is being forwarded in the event that the memory of neuron at $t$\textsuperscript{th} STS does not cross the threshold $T_h$. $\beta$ and $T_h$ can be hand-tuned or may be treated as trainable parameters at the time of training. Since there exists a discontinuity in the dynamics of spiking neurons, backpropagation is not deployed in its vanilla form; instead, surrogate backpropagation \cite{8891809,garg2023neuroscienceinspiredscientificmachine1} is used. In surrogate backpropagation, during the backward pass of the deep learning model, the discontinuity observed in VSN's dynamics is approximated using an equivalent continuous function. This assists in computing the gradients, which are then used during optimization. It should be noted that no change in neuron dynamics is carried out during the forward pass of the deep learning model. In the following numerical illustrations, we use the fast-sigmoid function \cite{zenke2018superspike} as our surrogate function. Now, since the energy consumption of any deep learning model is affected by the computations involved in the model, a neuron's firing activity or spiking activity becomes a good measure of energy consumption. Variants of the same are used in the literature \cite{kahana2022spiking,garg2024neuroscience} to judge the efficacy of spiking neurons. In this manuscript, we use the spiking activity $S$ computed as,
\begin{equation}
    S = \dfrac{\text{Number of spikes produced in spiking layer $L_s$}}{\text{Total number of possible spikes in spiking layer $L_s$}},
\end{equation}
to show the efficacy of the VSNs within the graph architectures.
\begin{figure}[ht!]
    \centering
    \includegraphics[width=0.9\linewidth]{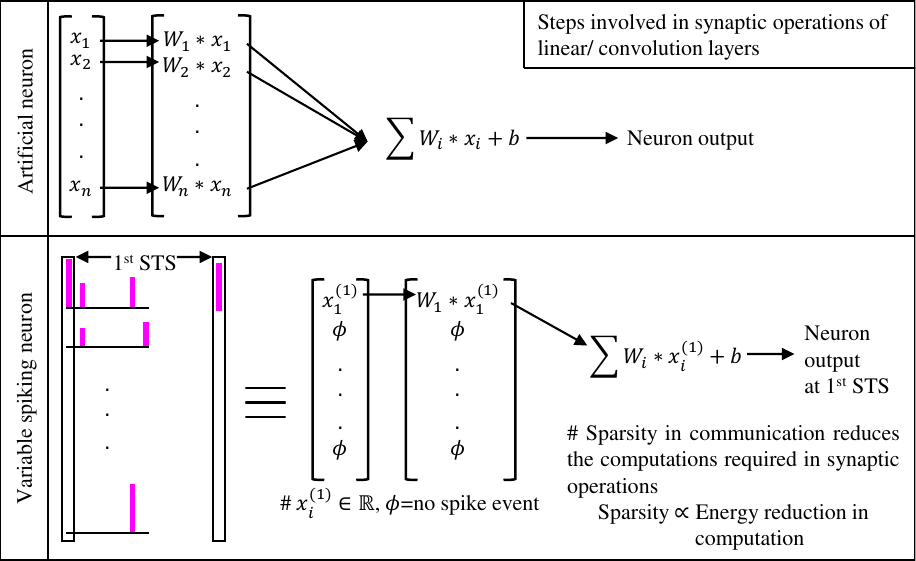}
    \caption{Schematic to draw a parallel between spiking activity and computations involved (in synaptic operations) in deep learning models.}
    \label{fig: spk act}
\end{figure}
A schematic is shown in Fig. \ref{fig: spk act} to draw a parallel between spiking activity and computations involved (in synaptic operations). The manuscript \cite{garg2024neuroscience,garg2023neuroscienceinspiredscientificmachine1} suggests that if the spiking activity of VSNs is less than one or 100\%, they prove to be more energy efficient in performing synaptic operations of convolution layers in the deep learning model. Now to further curb communication between neurons, we can use a loss function that restricts the spiking activity of VSN. This loss $L_{slf}$ is termed the Spiking Loss Function (SLF) and is computed as,
\begin{equation}
    L_{slf} = \alpha_LL_v + \beta_LS,
\end{equation}
where $S$ is the spiking activity or firing rate of the neuron, and $L_v$ is the vanilla loss function. $\alpha_L$ and $\beta_L$ are weights assigned to each component of the loss function, and these can be tuned as hyper-parameters in order to strike a good balance between accuracy and sparsity of communication.

\begin{figure}[ht!]
    \centering
    \includegraphics[width=0.9\linewidth]{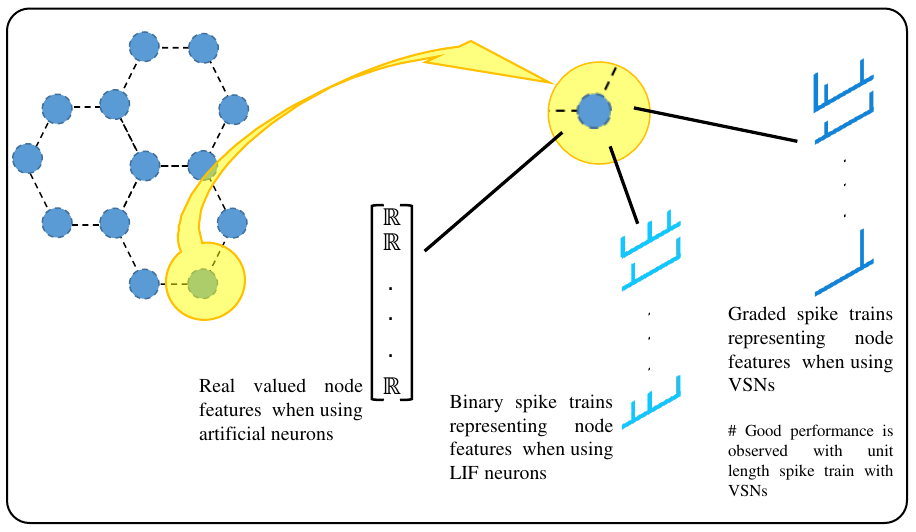}
    \caption{Schematic showing the quality of node features in different variants of GNNs.}
    \label{fig: feat quality}
\end{figure}

The VSNs, as discussed replace the continuous activation functions of a deep learning model. To convert a deep learning model into a spiking deep learning model, all activations need to be replaced by spiking neurons. However, this may lead to excessive sparsity in communication and information flow may stop entirely. This makes training inefficient and subsequently increases error in model's predictions. Hence, when converting GNNs to spiking GNNs, we propose replacing only few activations with spiking neurons. If some activations in the vanilla GNN network are retained as is and only the remaining are replaced by VSNs, the resulting architecture is termed HVS-GNNs. This selective replacement of continuous activations is essential to strike a good balance between accuracy and energy efficiency. Based on our experimentation, we found that replacing activations in between two successive GNN layers yield good results and makes the transition easier to implement. A schematic for the quality of node features in various variants of GNNs can be seen in Fig. \ref{fig: feat quality}.
\section{Numerical Illustrations}\label{section: ni}
To test the efficacy of the proposed HVS-GNNs, three examples from computational solid mechanics are presented in this section. In the first and second examples, the mechanical properties of polycrystalline materials are learned. Specifically, the stiffness and yield strength of $\alpha$-titanium polycrystal having different crystallographic structures are learned in the first example. The magnetostriction property for Terfenol-D alloy with different microstructures and under the influence of an external magnetic field is learned in the second example. Lastly, in the third example, the tensile stress field of a porous graphene membrane with atomic structural defects is learned. The first and second examples produce scalar outputs corresponding to graph inputs, whereas the third example learns the mapping between graph inputs and graph outputs. 
For these examples, we use HVS-GNNs with different arrangements of VSNs. For benchmarking and comparative study, we also train vanilla GNNs utilizing artificial neurons (A-GNNs) and Hybrid GNNs utilizing LIF neurons (HLIF-GNNs). 
Table \ref{tab:nomenclature} explains the nomenclature used to describe various deep learning models in the following case studies.
\begin{table}[ht!]
    \centering
    \caption{Nomenclature used to describe different variants of deep learning architectures in following examples.}
    \vspace{0.5em}
    \label{tab:nomenclature}
    \begin{tabular}{p{0.25\linewidth}p{0.65\linewidth}}
     \toprule
     Symbol & Description \\
     \midrule
     A-GNN & Graph neural network with vanilla activations in the architecture of the given example.\\
     HVS-GNN\textsuperscript{\#} & Graph neural networks with VSNs in positions as per version \# of the architecture in the given example.\\
     HVS-GNN\textsuperscript{\#}-SLF & HVS-GNN\textsuperscript{\#} networks trained using SLF to constrain spiking activity.\\
     HLIF-GNN\textsuperscript{\#} & Graph neural networks with LIF neurons in positions as per version \# of the architecture in the given example.
     \\\bottomrule
    \end{tabular}
\end{table}
The hybrid spiking GNNs used in this paper deploy spiking neurons between two successive layers, the activations within a message-passing layer or a recurrent cell are kept as is. Details for dataset and the network architecture adopted in a various examples are included in their respective sub-sections. 
\subsection{Example 1: Learning mechanical properties of $\alpha$-titanium polycrystals}
\begin{figure}[ht!]
    \centering
    \includegraphics[width=0.95\linewidth]{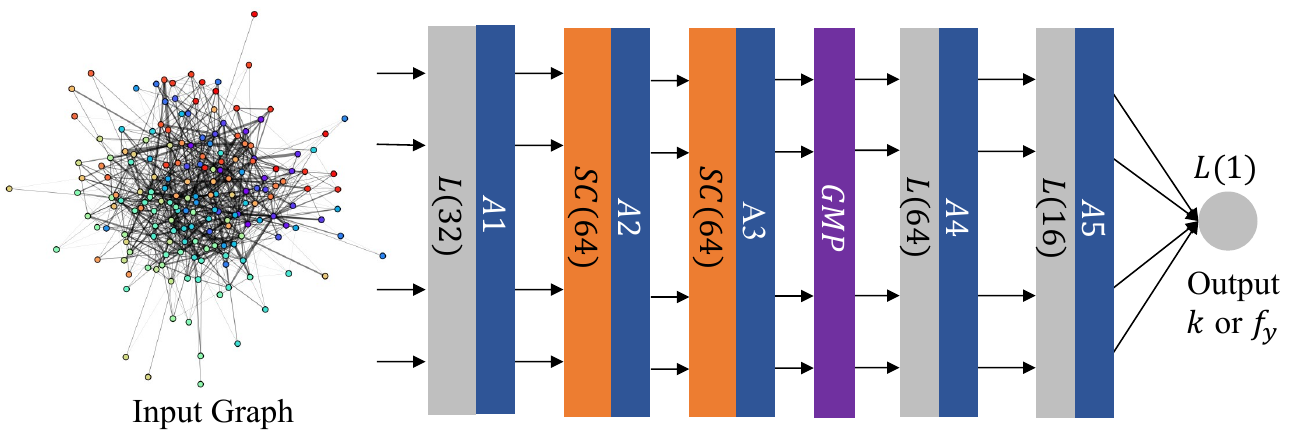}
    \caption{Schematic for the base deep learning architecture used in Example-1.}
    \label{fig: arch 1}
\end{figure}

In the first example, the objective is to learn the mechanical properties of $\alpha$-titanium polycrystals corresponding to their crystallographic textures \cite{HESTROFFER2023}.
Specifically, the stiffness $k$ and yield strength $f_y$ are learned, given the crystals' Microstructure Volume Elements (MVEs). 
The dataset contains 1200 MVEs, stochastically generated using 12 distinct crystallographic textures labeled from A to L and their corresponding mechanical properties, $k$ and $f_y$.
The MVEs are represented as graphs where nodes are individual grains and edges connect grains that share a boundary. Each node has a $5\times 1$ feature vector, the first four elements of which represent the grain-average crystallographic orientation reduced to the hexagonal fundamental region, and the fifth element represents the grain size. 
%
%
%
The architecture of the deep learning model adopted here is as follows \cite{HESTROFFER2023},
\begin{equation*}
    L(32) \rightarrow A1 \rightarrow SC(64) \rightarrow A2 \rightarrow SC(64) \rightarrow A3 \rightarrow GMP \rightarrow L(64) \rightarrow A4 \rightarrow L(16) \rightarrow A5 \rightarrow L(1),
\end{equation*}
where $L(\#)$ represents a linear layer with $\#$ number of layer nodes and $A\#$ represents the $\#$\textsuperscript{th} activation layer. The $SC(\#)$ layer is a message-passing SAGEConv \cite{hamilton2017inductive,HESTROFFER2023} layer with $\#$ output channels. For the A-GNN model in the current example, all activation layers in the above network are replaced by ReLU activation. For variation 1 of spiking graph networks in the current example, i.e., HVS-GNN\textsuperscript{1} and  HLIF-GNN\textsuperscript{1}, activations A1, A2, and A3 are replaced with respective spiking neurons, and in variation 2, i.e., HVS-GNN\textsuperscript{2} and  HLIF-GNN\textsuperscript{2}, activations A1-A5 are replaced by spiking neurons. A schematic for the base deep learning architecture is shown in Fig. \ref{fig: arch 1}.
All networks are trained using ADAM optimizer with 0.001 learning rate, and the training was done for 600 epochs.

Two sets of networks are trained to give the stiffness and yield strength of a crystal, respectively, as outputs. For training the deep learning models in the current example, a training dataset containing 630 MVEs corresponding to textures A-G (90 from each) is used. While testing, the following two cases are taken: (i) Evaluation-1: testing on a dataset containing 70 MVEs derived from textures A-G (10 from each), i.e., within distribution prediction, and (ii) Evaluation-2: testing on a dataset containing 500 MVEs derived from textures H-L (100 from each), i.e., out of distribution prediction.

\begin{table}[ht!]
  \centering
  \caption{MSE values observed when comparing model predictions and ground truth for Evaluation-1 (training and testing on datasets corresponding to A-G textures) and  Evaluation-2 (training on datasets corresponding to A-G textures and predicting on datasets corresponding to textures H-L) in Example 1.}
  \vspace{0.5em}
  \label{tab:ex1MSE}
  \begin{tabular}{lcccc}
    \toprule
    MSE ($\times 10^{-3}$)& \multicolumn{2}{c}{Evaluation-1}& \multicolumn{2}{c}{Evaluation-2}\\
    \cmidrule(lr){2-3}\cmidrule(r){4-5}
    Model& $k$ & $f_y$ &  $k$ & $f_y$\\
    \midrule
    A-GNN & {2.11} & {10.06} & {6.69} & {33.42}\\
    \hdashline
    HLIF-GNN\textsuperscript{1} & 5.34 & 13.68 & 256.62 & 139.27\\
    HVS-GNN\textsuperscript{1} & \textbf{1.79} & 7.83 & 49.01 & 47.51\\
    HLIF-GNN\textsuperscript{2} & 3.72 & 10.72 & 25.67 & 130.45\\
    HVS-GNN\textsuperscript{2} & 2.48 & \textbf{6.93} & \textbf{10.00} & \textbf{31.01}\\
    \bottomrule
  \end{tabular}
\end{table}
Table \ref{tab:ex1MSE} shows the Mean Square Errors (MSEs) obtained when comparing true values and network predictions corresponding to the test datasets of the two evaluations. It can be observed that the performance of HVS-GNN models is better than the HLIF-GNN models for both variants of network architectures, and despite sparse communication, the performance of HVS-GNN\textsuperscript{2} is comparable to the performance of A-GNN. It should be noted that the performance of HVS-GNN\textsuperscript{1} in evaluation 1, which is within distribution prediction, is marginally better than that of the A-GNN when predicting stiffness $k$.
\begin{figure}[ht!]
    \centering
    \includegraphics[width=0.95\textwidth]{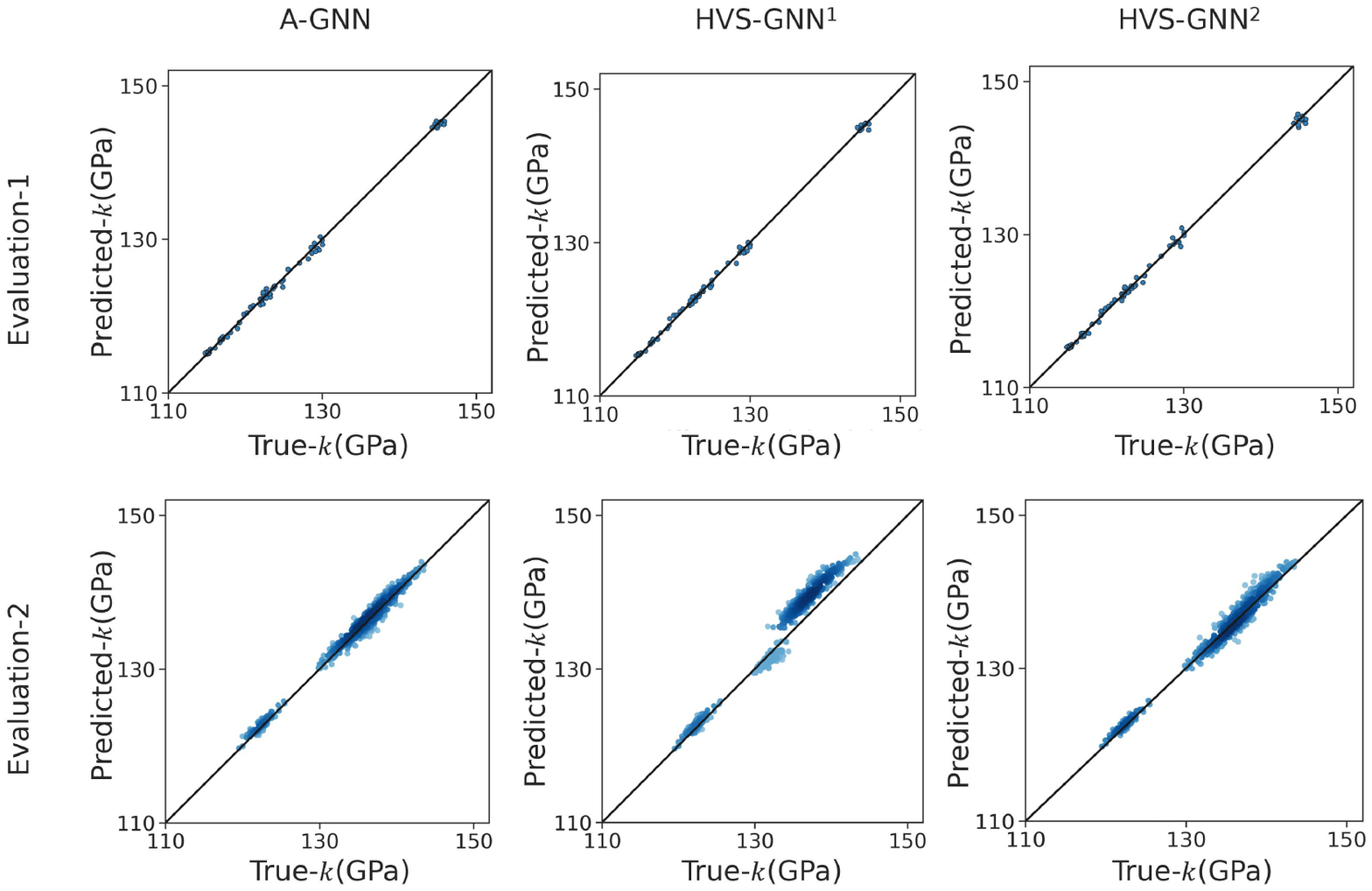}
    \caption{Plots for predicted vs true stiffness $k$ for both prediction evaluation cases in the first example.}
    \label{fig:plots_stiff}
\end{figure}
\begin{figure}[ht!]
    \centering
    \includegraphics[width=0.95\textwidth]{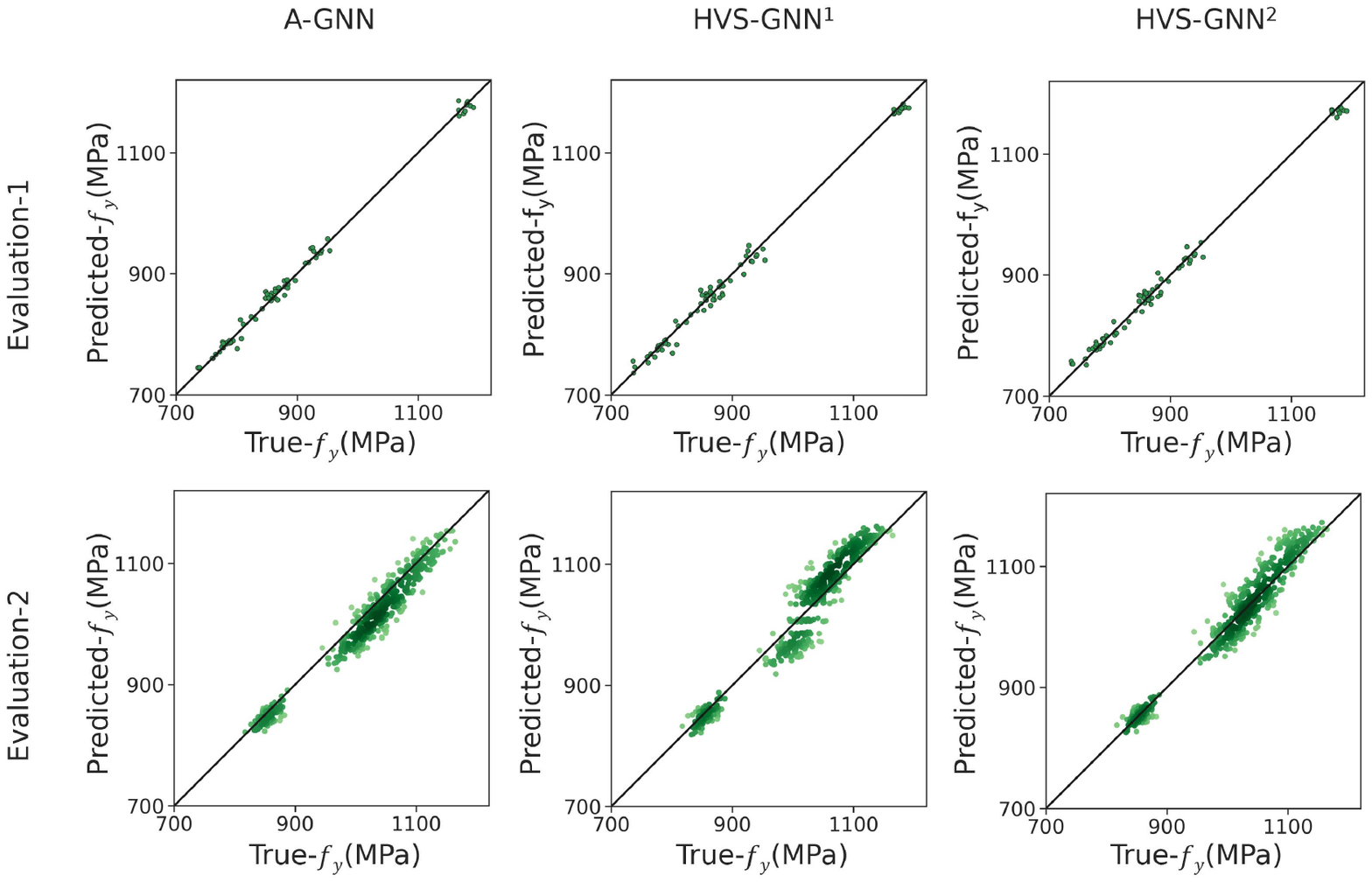}
    \caption{Plots for predicted vs true yield strength $f_y$ for both prediction evaluation cases in the first example.}
    \label{fig:plots_strength}
\end{figure}
Figs. \ref{fig:plots_stiff} and \ref{fig:plots_strength} show the predictions vs ground truth plots for stiffness $k$ and yield strength $f_y$ respectively. The same reaffirms the trends observed in Table \ref{tab:ex1MSE}.

\begin{figure}[ht!]%
    \centering
    \subfloat[\centering HVS-GNN$^{1}$ and HVS-GNN$^{1}$-SLF networks]{{\includegraphics[width=0.42\textwidth]{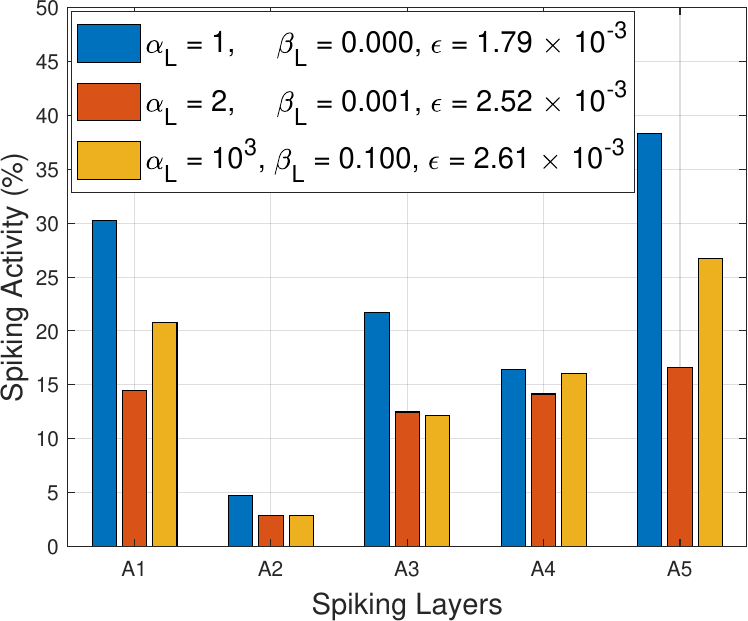} }}%
    \qquad
    \subfloat[\centering HVS-GNN$^{2}$ and HVS-GNN$^{2}$-SLF networks]{{\includegraphics[width=0.42\textwidth]{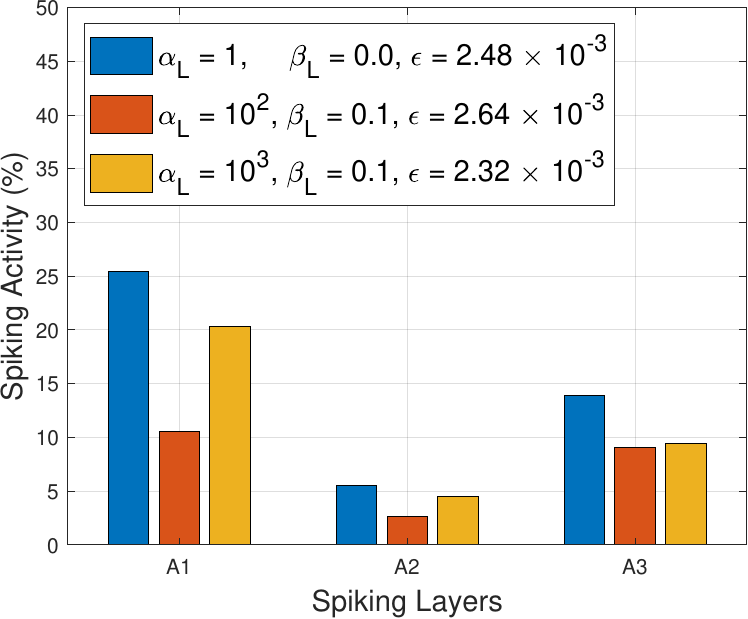} }}%
    \caption{Percentage spiking activity observed in various spiking layers when predicting the stiffness $k$ for textures A-G in the first example. $\alpha$ and $\beta$ are the weights of the SLF and $\beta = 0$ represents the case when the vanilla loss function is used. $\epsilon$ represents the observed MSE values when comparing the predictions against the ground truth.}
    \label{fig:spk_stiff1}%
\end{figure}
\begin{figure}[ht!]%
    \centering
    \subfloat[\centering HVS-GNN$^{1}$ and HVS-GNN$^{1}$-SLF networks]{{\includegraphics[width=0.42\textwidth]{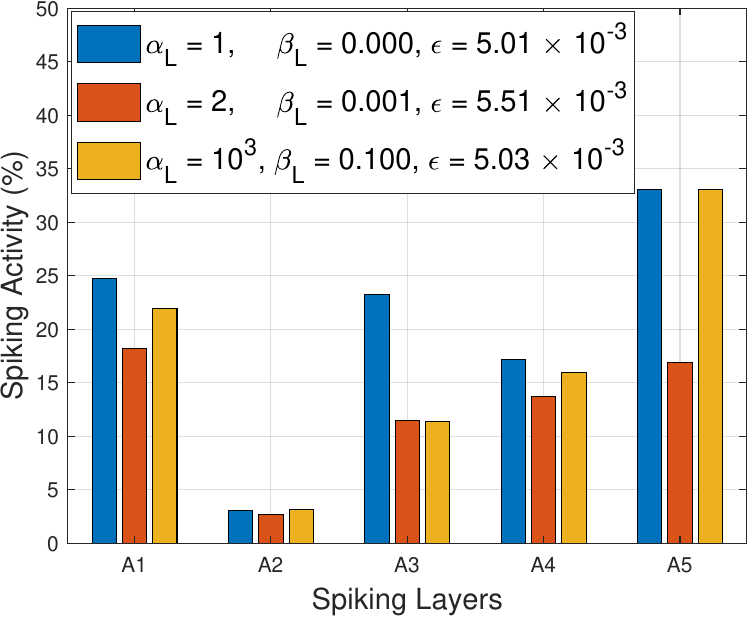} }}%
    \qquad
    \subfloat[\centering HVS-GNN$^{2}$ and HVS-GNN$^{2}$-SLF networks]{{\includegraphics[width=0.42\textwidth]{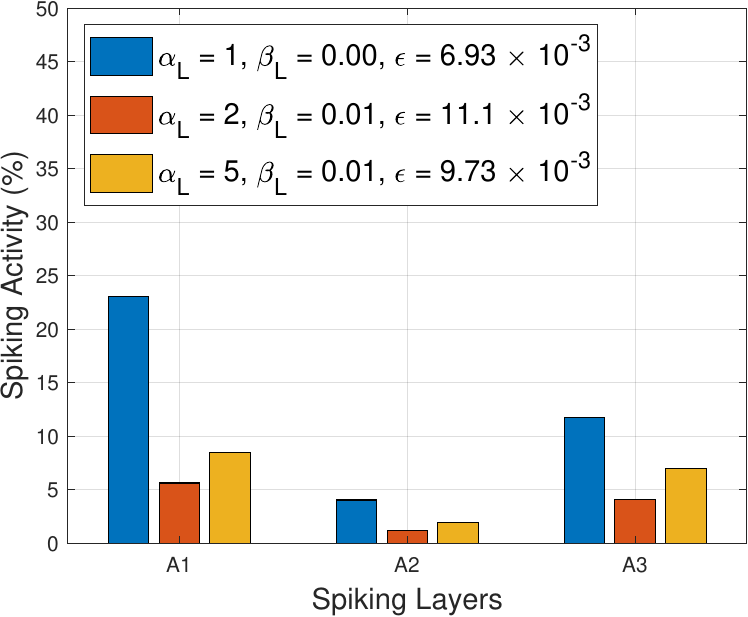} }}%
    \caption{Percentage spiking activity observed in various spiking layers when predicting the yield strength $f_y$ for textures A-G in the first example. $\alpha$ and $\beta$ are the weights of the SLF and $\beta = 0$ represents the case when the vanilla loss function is used. $\epsilon$ represents the observed MSE values when comparing the predictions against the ground truth.}%
    \label{fig:spk_strength1}%
\end{figure}
Figs. \ref{fig:spk_stiff1} and \ref{fig:spk_strength1} show the spiking activity observed in the VSN layers of HVS-GNN models when predicting $k$ and $f_y$ respectively using evaluation-1 datasets. They show a drop in spiking activity when using the SLF function, and as can be observed, the increase in MSE after using SLF is only marginal, and the same is still comparable to the values observed when using the A-GNN model.
\subsection{Example 2: Prediction of magnetostriction in polycrystalline material (Terfenol-D)}
In this example, we learn the magnetostriction property of Terfenol-D alloy corresponding to its polycrystalline microstructure, and external magnetic field. Magnetostriction refers to the ability of microstructure to undergo mechanical strain or an observable change in shape or dimension due to the effect of a magnetic field. The dataset in this example comprises 3D polycrystalline microstructure generated using Dream.3D software \cite{groeber2014dream}. Each microstructure contains a different number of grains, ranging from 12 to 297 grains.
The polycrystalline microstructures are represented as graphs, where each grain is a node, and the edges represent the grain boundaries shared between the grains. The features of each node have 5 components, comprising three Euler angles describing the grain orientation, grain size depicting the number of voxels occupied by the grain, and the number of neighboring grains. 
Different graph samples are padded to maintain a consistent number of nodes across all samples. Apart from the graphs representing polycrystalline microstructure, the external magnetic field $\bm H_x$ is also used as input. The output is the effective magnetostriction $\lambda^{eff}_{xx}$ along the x-axis, given the polycrystalline microstructure and the external magnetic field. 
%
The architecture of the deep learning model adopted in the current example is as follows \cite{dai2021graph},
\begin{multline}    
    MP1 \rightarrow L(40) \rightarrow A1 \rightarrow MP2 \rightarrow L(3) \rightarrow A2 \rightarrow \\ [\mathcal F+\mathcal C] \rightarrow L(1024)\rightarrow A3 \rightarrow L(128) \rightarrow A4 \rightarrow L(1),
\end{multline}
where the $MP\#$ layer is a message-passing layer with \# representing the $\#$\textsuperscript{th} message-passing layer. The first message passing layer, $MP1$, accepts the graph input. $[\mathcal F+\mathcal C]$ refers to the flattening and concatenation operations that follow after $A2$. The flattening layer flattens the graph feature coming from activation $A2$, arranging them in a vector. It then concatenates this flattened feature vector with $\bm H_x$ to introduce the effect of the external magnetic field of the current sample in the model. This concatenated output is then passed on to the subsequent linear layer $L(1024)$.

\begin{figure}[ht!]
    \centering
    \includegraphics[width=0.95\linewidth]{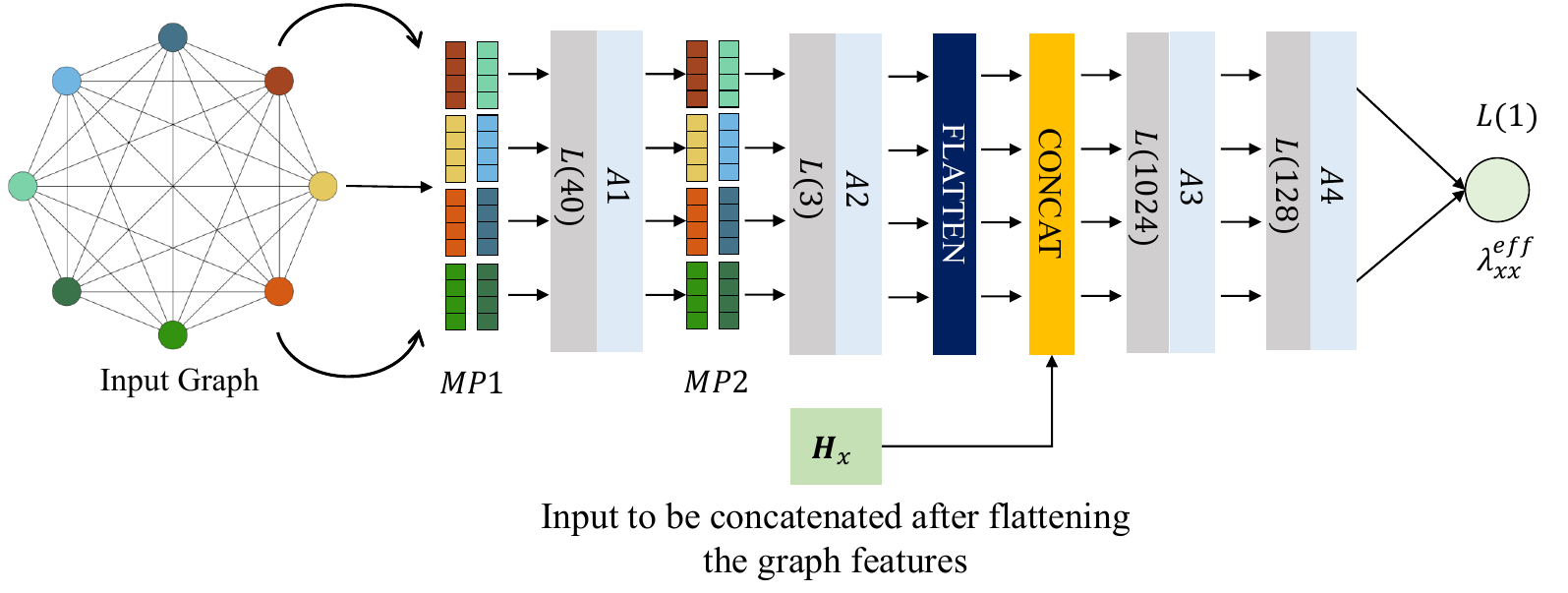}
    \caption{Schematic for the base deep learning architecture used in Example-2. The input graph sample shown is a placeholder and does not reflect the actual graph of the dataset.}
    \label{fig: arch 2}
\end{figure}

For the A-GNN model in the current example, the $A1$ and $A2$ activation layers in the above network are replaced by sigmoid activation, and the $A3$ and $A4$ are replaced by ReLU activation. For variation 1 of spiking graph networks in the current example, i.e., HVS-GNN\textsuperscript{1} and  HLIF-GNN\textsuperscript{1}, activations $A3$ and $A4$ are replaced with respective spiking neurons, and in variation 2, i.e., HVS-GNN\textsuperscript{2} and  HLIF-GNN\textsuperscript{2}, activations $A1$ to $A4$ are replaced by spiking neurons. A schematic for the base deep learning architecture is shown in Fig. \ref{fig: arch 2}.
All networks are trained using the ADAM optimizer with a learning rate of 0.0005, and training was done for 500 epochs.
We have a total of 2,287 samples generated using 492 unique polycrystalline microstructures, with each microstructure associated with four to five different external magnetic fields. A total of 1,831 samples are used for training,
228 data points for validation, 
and 228 samples are used for testing.

\begin{table}[ht!]
  \centering
  \caption{MSE values observed when comparing model predictions and ground truth in Example 2.}
  \vspace{0.5em}
  \label{tab:mytable_mag}
    \begin{tabular}{c:c:cccc}
    \toprule
     Model & A-GNN & HLIF-GNN\textsuperscript{1} & HVS-GNN\textsuperscript{1} & HLIF-GNN\textsuperscript{2} & HVS-GNN\textsuperscript{2}\\
     \midrule
     MSE ($\times 10^{-9})$ & {9.67} & 9.88 & 8.77 & 7.08 & \textbf{4.29} \\
    \bottomrule
  \end{tabular}
\end{table}

\begin{figure}[ht!]
    \centering
    \includegraphics[width=\textwidth]{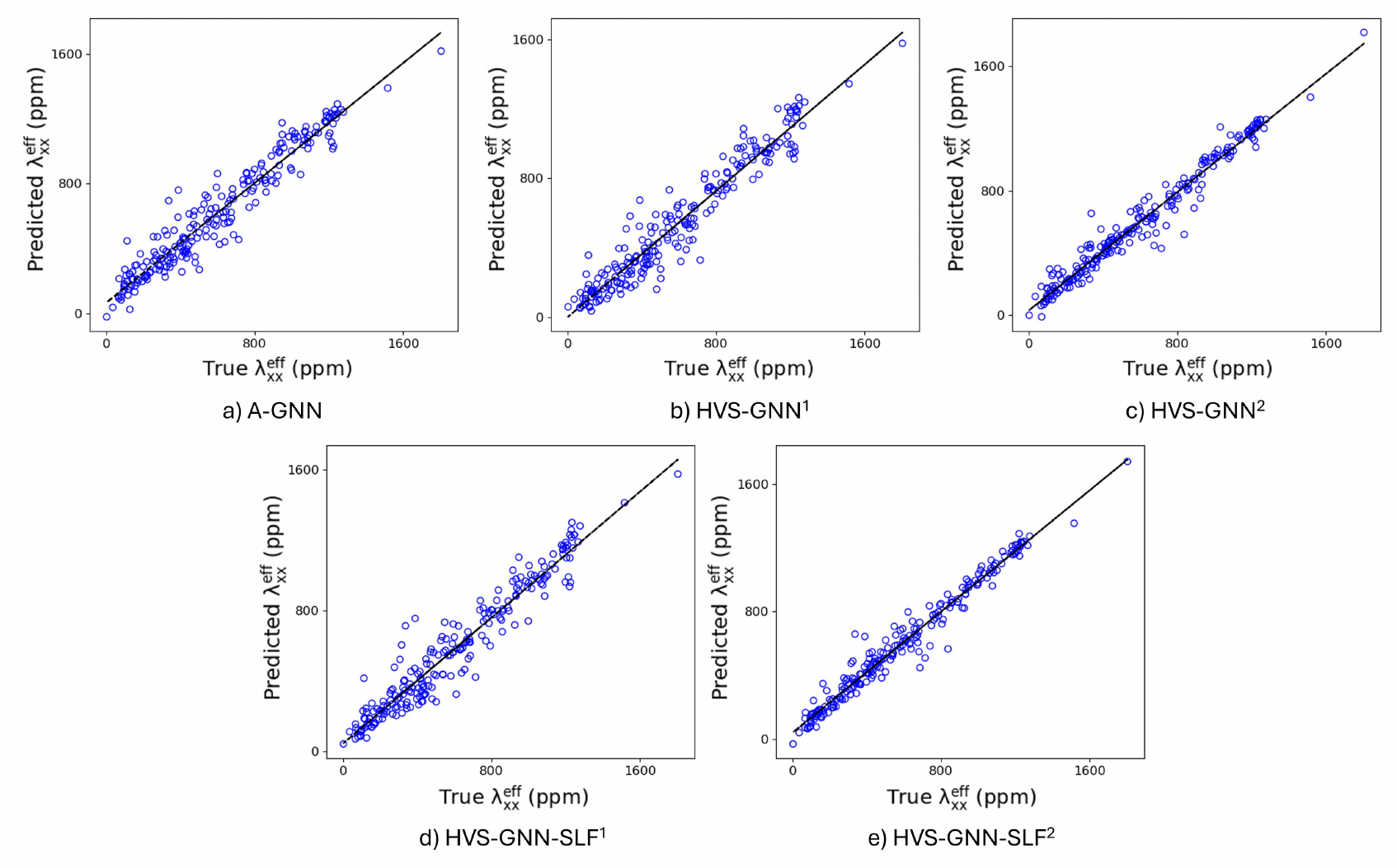}
    \caption{Plots for predicted vs true $\lambda^{eff}_{xx}$ in the second example.}
    \label{fig:plots_mag}
\end{figure}
Table \ref{tab:mytable_mag} shows the MSE errors obtained when comparing ground truth against the trained model predictions. As can be seen, in the current example, the performance of HVS-GNN\textsuperscript{1} is comparable to the performance of A-GNN, and the performance of HVS-GNN\textsuperscript{2} is better than the performance of all other networks. Fig. \ref{fig:plots_mag} shows the ground truth plotted against the model predictions for A-GNN and HVS-GNN networks. It also shows plots for HVS-GNN, when the same are used with SLF to reduce spiking activity of neurons. As can be seen for HVS-GNN\textsuperscript{2} network, the model predictions are closer to true values, a trend similar to what was observed in Table \ref{tab:mytable_mag}. Furthermore, even after using SLF, the model predictions for HVS-GNN\textsuperscript{2} network, closely follow the ground truth.

\begin{figure}[ht!]
    \centering
    \includegraphics[width=0.475\textwidth]{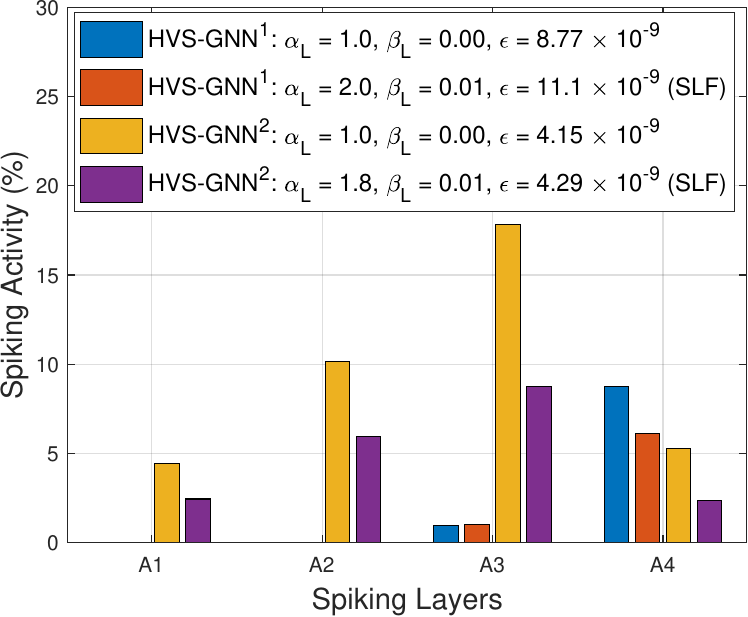}
    \caption{Percentage spiking activity observed in various spiking layers when predicting $\lambda^{eff}_{xx}$ in the second example. $\alpha$ and $\beta$ are the weights of the SLF and $\beta = 0$ represents the case when the vanilla loss function is used. $\epsilon$ represents the observed MSE values when comparing the predictions against the ground truth.}
    \label{fig:spk_mag}
\end{figure}
Fig. \ref{fig:spk_mag} shows the average spiking activity of VSN layers used in HVS-GNN networks. A trend similar to that in the previous example is observed, where we observe a drop in spiking activity when using SLF, whereas the increase in error observed is only marginal and is still comparable to that observed when using A-GNN. In this particular example, the performance on HVS-GNN\textsuperscript{2} is better than A-GNN, and the same remains true when SLF is used to train the HVS-GNN\textsuperscript{2} network. 
\subsection{Example 3: Prediction of the tensile stress field in a graphene memberane}
In the third example, we learn the tensile stress field in porous graphene membranes. 
The dataset consists of 2000 crystals of porous graphene membranes of size 127.9\r{A} $\times$ 127.8\r{A} with atomic structural defects and their corresponding tensile stress field observed when stretched under a strain of 5\%. The crystals are represented using graphs, and a node represents each atom of a crystal. An edge connects two nodes if the distance between corresponding atoms is less than a cut-off distance, set as 1.92\r{A}. Node features for input are the coordinates of each atom in the two-dimensional space and for output, the node feature is the tensile stress value at that node. 

\begin{figure}[ht!]
    \centering
    \includegraphics[width=\linewidth]{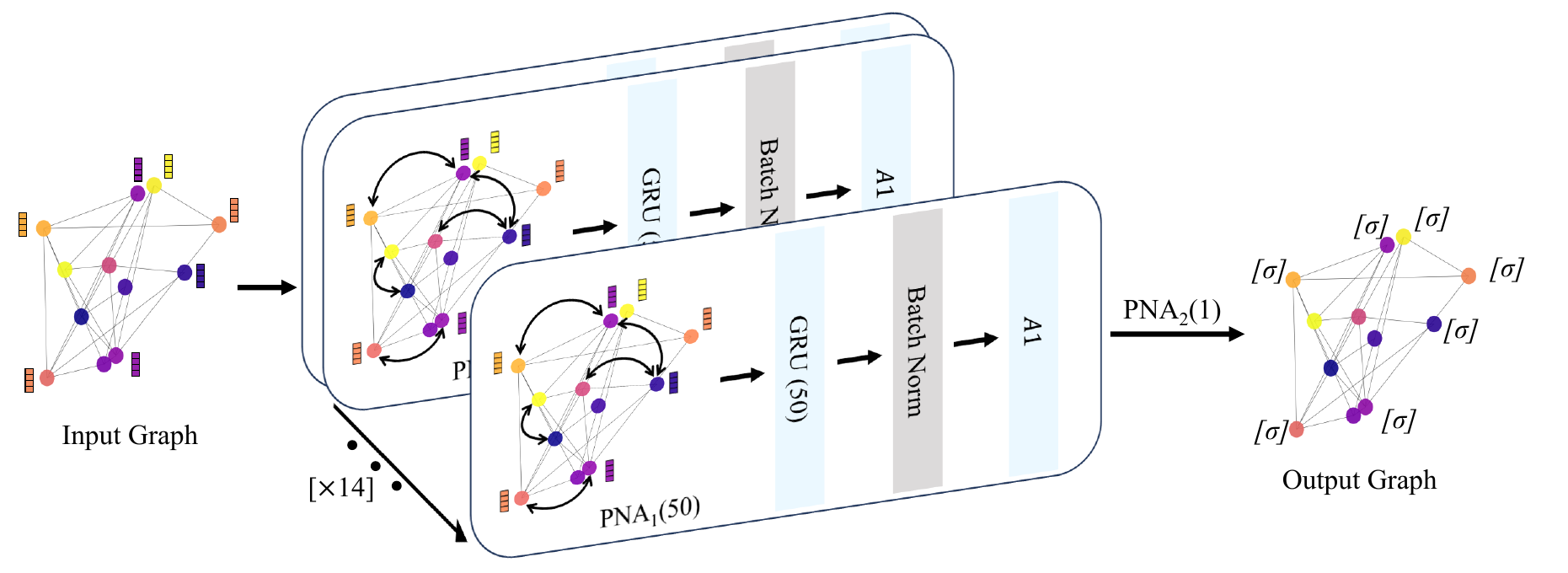}
    \caption{Schematic for the base deep learning architecture used in Example-3. The input graph sample shown is a placeholder and does not reflect the actual graph of the dataset.}
    \label{fig: arch 3}
\end{figure}

The base deep learning model architecture and the dataset for this example are taken from \cite{yang2022linking}. The architecture of the deep learning model adopted here is as follows,
\begin{equation*}
    [\,\text{PNA}_1(50) \rightarrow \text{GRU}(50) \rightarrow \text{BatchNorm} \rightarrow A1\,] \times 14 \rightarrow \text{PNA}_2(1),
\end{equation*}
where PNA\textsubscript{i}(\#) is a PNAConv layer \cite{corso2020principal,yang2022linking} with \# output channels, and it acts as a message-passing layer. 
The PNA\textsubscript{2}(1) at the end is used for as a readout layer. GRU(\#) is the GRUCell function \cite{chung2014empiricalevaluationgatedrecurrent} where \# represents the number of output channels. $[\cdot]\times 14$ in the above architecture denotes that the block of various layers [$\cdot$] is repeated sequentially 14 times within the architecture. For the A-GNN model in the current example, the A1 activation layer in the above network is replaced by ReLU activation. For variation 1 of spiking graph networks in the current example, the output of PNA\textsubscript{1}(50) is wrapped by the spiking neuron layer, and also, the activation A1 is replaced by the spiking neuron layer. The same is done for each of the 14 blocks of the architecture. In variation 2, only the A1 activation in each of the 14 blocks is replaced by spiking neurons. A schematic for the base deep learning architecture is shown in Fig. \ref{fig: arch 3}.
All networks are trained using the ADAM optimizer with a learning rate of 0.001, and the training was done for 240 epochs. Out of the total of 2000 crystals in the dataset, 1400 have been used for training, 50 for validation, and 550 for testing.

\begin{table}[ht!]
  \centering
  \caption{MSE values observed when comparing model predictions and ground truth in the third example.}
  \vspace{0.5em}
  \label{tab:mytable_atm}
  \begin{tabular}{c:c:ccc}
    \toprule
     Model & A-GNN & HLIF-GNN\textsuperscript{1} & HVS-GNN\textsuperscript{1} & HVS-GNN\textsuperscript{2}
     \\\midrule
     MSE & \textbf{0.535} & 1.334 & 0.645 & \textbf{0.576} \\
    \bottomrule
  \end{tabular}
\end{table}
\begin{figure}[ht!]
    \centering
    \includegraphics[width=\textwidth]{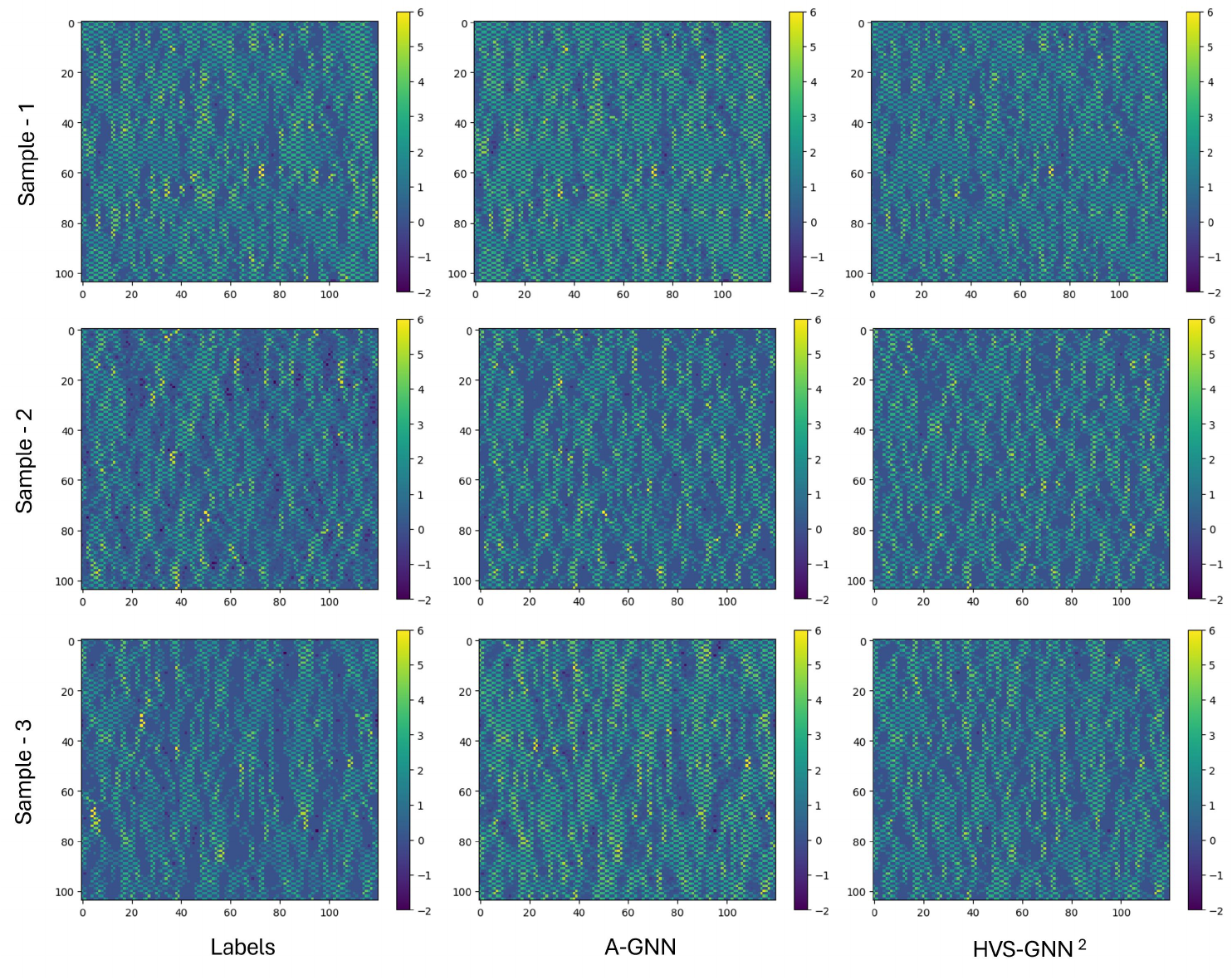}
    \caption{Model predictions compared against the ground truth (Labels) for three samples from the test dataset in the third example.}
    \label{fig:atomic_maps}
\end{figure}
Table \ref{tab:mytable_atm} shows the MSE values obtained when comparing model predictions against the ground truth and Fig. \ref{fig:atomic_maps} plots the model predictions and the ground truth corresponding to one sample from the test dataset. The trend observed in the previous examples continues here, and as can be observed, the performance of HVS-GNN models is better than the performance of HLIF-GNN models and is comparable to the performance of the A-GNN model.

\begin{figure}[ht!]%
    \centering
    \subfloat[\centering Activation around PNA\textsubscript{1}(50) layer.]{{\includegraphics[width=0.45\textwidth]{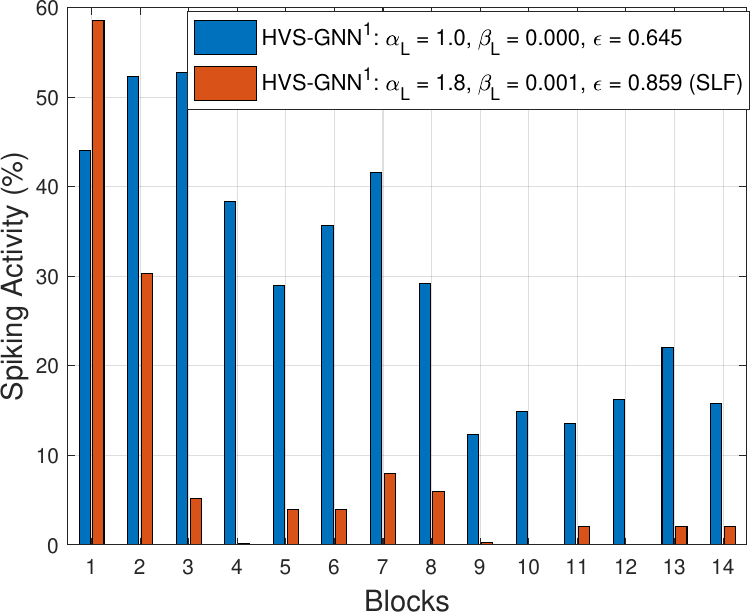} }}%
    \qquad
    \subfloat[\centering Layer $A1$.]{{\includegraphics[width=0.45\textwidth]{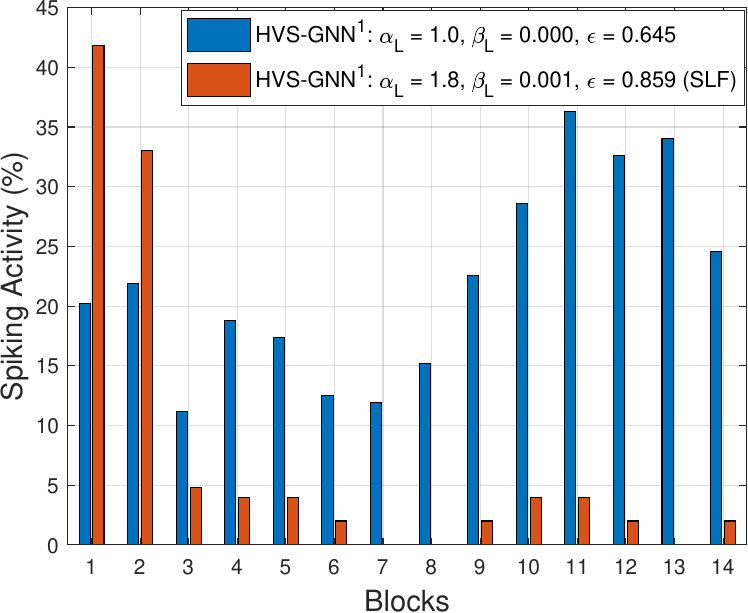} }}%
    \caption{Percentage spiking activity plots for various layers in all 14 blocks of HVS-GNN\textsuperscript{1} architecture. $\alpha$ and $\beta$ are the weights of the SLF and $\beta = 0$ represents the case when the vanilla loss function is used. $\epsilon$ represents the observed MSE values when comparing the predictions against the ground truth.}%
    \label{fig:spk_afield_vsn1}%
\end{figure}
\begin{figure}[ht!]
    \centering
    \includegraphics[width= 0.45\textwidth]{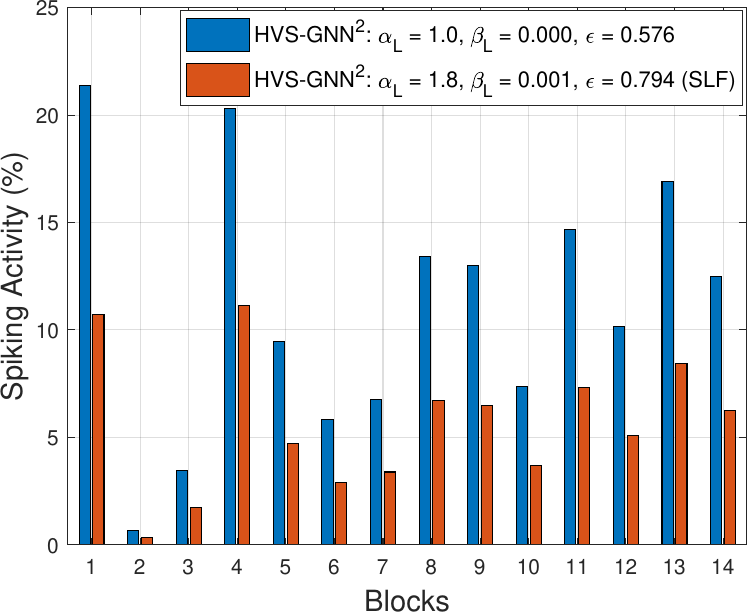}
    \caption{Percentage spiking activity for $A1$ layer in all 14 blocks of HVS-GNN\textsuperscript{2} architecture. $\alpha$ and $\beta$ are the weights of the SLF and $\beta = 0$ represents the case when the vanilla loss function is used. $\epsilon$ represents the observed MSE values when comparing the predictions against the ground truth.}
\label{fig:spk_afield_vsn2}
\end{figure}
Fig. \ref{fig:spk_afield_vsn2} shows the spiking activity in VSN layers of HVS-GNN models and, similar to the previous examples, a decrease in spiking activity is observed when using SLF while training the models. The increased MSE when using SLF to train HVS-GNN models, despite sparse communication, is less than when LIF-GNN is used for prediction. 
\section{Conclusion}\label{section: conclusion}
GNNs represent a branch of deep learning architectures that utilize techniques like message passing to analyze graph based datasets effectively. This makes them useful in fields like computational mechanics, where complex datasets often arise from irregular grids or the analysis of individual particles within a structure. However, as the complexity of the problem and the deep learning model increases, the energy budget associated with the GNN models utilizing artificial neurons (A-GNN) and, thus, continuous activations increases tremendously. This is because of increased computations in the model, which makes such models impractical for applications where the overall energy resource available is limited. 
In response to this challenge, we propose HVS-GNNs, a hybrid spiking variant of vanilla GNNs. HVS-GNNs utilize VSNs that promote sparse communication, decreasing the computational load and ultimately reducing the energy budget. The VSNs replace the activation functions of A-GNN models; however, they need not replace all the activations, thus giving rise to HVS-GNN architectures. We test the HVS-GNN models for three different examples in computational solid mechanics. The selected examples involve both scalar outputs and graph outputs corresponding to graph inputs. The findings of the paper are as follows,
\begin{itemize}
    \item \textbf{Regression performance - Graph datasets}: The performance of HVS-GNNs in their intended use case, i.e., for regression tasks, is at par with the performance of the current state-of-the-art A-GNN model. In fact, in the second example involving magnetostriction prediction in Terfenol-D, the HVS-GNN model surpassed the accuracy achieved when using the A-GNN model.
    \item \textbf{Comparison against LIF neurons}: For all the three examples, HVS-GNN's performance in regression tasks is decidedly better than the performance of HLIF-GNN models. In example 2, HLIF-GNN's performance is at par with the A-GNN model; however, in that example, the HVS-GNN models outperform both HLIF-GNN and A-GNN models.
    \item \textbf{Spiking Activity:} As discussed previously, the goal of using spiking neurons is to increase sparsity in communication, leading to reduced computations (when using event-driven hardware) and hence reduced energy budget. In all three examples, the spiking activity of VSN layers is well below 100\%, going to a maximum of ~37\%, ~18\%, and ~59\% in examples 1, 2, and 3, respectively. Furthermore, after using SLF during training, we observe a reduction in spiking activity in several, if not all, spiking layers of HVS-GNN models.
    
    
\end{itemize}
It was further observed from the results that the spiking activity of HVS-GNN models can be brought down significantly when using SLF during training at marginal cost to overall accuracy. The authors would like to note here that the energy conservation capacity of VSNs is only maximized when all activations are replaced by the VSNs. Also, a more comprehensive study is required to gauge the energy savings in a real-world scenario when running the HVS-GNNs on neuromorphic hardware. Overall, it can be concluded that the VSNs work well within the GNN architectures, and good performance can be expected out of the overall HVS-GNN models. 
\section*{Acknowledgment}
SG acknowledges the financial support received from the Ministry of Education, India, in the form of the Prime Minister's Research Fellows (PMRF) scholarship. SC acknowledges the financial support received from Anusandhan National Research Foundation (ANRF) via grant no. CRG/2023/007667 and from the Ministry of Port and Shipping via letter no. ST-14011/74/MT (356529). 

\appendix
\section{Leaky Integrate and Fire Neurons and artificial neurons}\label{appendix:LIF}
The dynamics of an artificial neuron is defined as,
\begin{equation}
    y = \sigma(z),
\end{equation}
where $z$ is the input to the neuron and for a linear node is computed as $z = wx+b$, $x$ being the output from a previous node and $w$ and $b$ representing the weights and bias, respectively. As can be observed, there is no mechanism to stop the flow of information within the dynamics of artificial neuron. This is where spiking neurons come into picture. To utilize the concept of spiking neurons in deep learning architectures, mathematical models of the biological neurons are required. The dynamics of a LIF neuron which strikes a good balance between biological plausibity and ease of implementation is defined as follows,
\begin{equation}
    \begin{aligned} 
    M^{(\bar{t})} & =\beta M^{(\bar{t}-1)}+z^{(\bar{t})} \\ y^{(\bar{t})} & =\left\{\begin{array}{ll}1 ; & M^{(\bar{t})} \geq T \\ 0 ; & M^{(\bar{t})}<T\end{array} \quad, \quad \text { if } y^{(\bar{t})}=1, M^{(\bar{t})} \leftarrow 0\right.,
    \end{aligned}
\end{equation}
where $M^{(\bar{t})}$ and $z^{(\bar{t})}$ represent the memory and input to LIF neuron, respectively, at $\bar{t}$ STS. The memory of a LIF neuron is multiplied by a leakage parameter $\beta$ after each STS. $y^{(\bar{t})}$ represents the output at $\bar{t}$ STS and is equal to one only if $M^{(\bar{t})}$ is greater then the threshold $T$ of neuron. Else no output is forwarded from the neuron. As can be observed here, the output at any STS can either be one or zero (representing no event). This limits the information-carrying capacity of a spike, and it has been observed that for regression tasks, the performance achieved when using binary spikes (as in the case of LIF neurons) is not at par with the performance achieved when using graded signals (as in the case of artificial neurons).
\end{document}